# Extracting Linguistic Speech Patterns of Japanese Fictional Characters Using Subword Units

Mika Kishino[1] and Kanako Komiya[2]

[1]Ibaraki University, Ibaraki, Japan
21nm722y@vc.ibaraki.ac.jp
[2]Tokyo University of Agriculture and Technology, Tokyo, Japan
kkomiya@go.tuat.ac.jp

***ABSTRACT***

*This study extracted and analyzed the linguistic speech patterns that characterize Japanese anime or game characters. Conventional morphological analyzers, such as MeCab, segment words with high performance, but they are unable to segment broken expressions or utterance endings that are not listed in the dictionary, which often appears in lines of anime or game characters. To overcome this challenge, we propose segmenting lines of Japanese anime or game characters using subword units that were proposed mainly for deep learning, and extracting frequently occurring strings to obtain expressions that characterize their utterances. We analyzed the subword units weighted by TF/IDF according to gender, age, and each anime character and show that they are linguistic speech patterns that are specific for each feature. Additionally, a classification experiment shows that the model with subword units outperformed that with the conventional method.*

## 1. Introduction

There is research in the field of natural language processing that focuses on linguistic styles and characterizes utterances of confined groups categorized by some features like gender or age. Japanese is a language whose expressions vary depending on gender, age, and relationships with dialog partners. In particular, Japanese anime and game characters sometimes speak with emphasis on character rather than reality. Furthermore, the way of talking of Japanese fictional characters is sometimes different from real people. For example, Funassyi, a Japanese mascot character, usually ends each utterance with "なっしー, nassyi" yet this ending is not found in a Japanese dictionary. Additionally, a cat character tends to add "にゃん, nyan", an onomatopoeia that expresses a cry of a cat at the end of each utterance. Human characters also have character-specific linguistic speech patterns in novels, anime, and games. They are known as role language [1] and it is related to characterization; the role language shows what role the speaker plays, and sometimes it is different from real conversation. For example, "僕, boku, I" is a first-person singular usually used for boys in novels, anime, and games, but it is also used for men and boys in real life. Therefore, in this study, we extracted and analyzed the linguistic speech patterns that characterize these characters using utterances of anime or game characters. In Japanese, morphological analysis is a basic technology for natural language processing because Japanese does not have word delimiters between words. Word segmentation and morphological analysis are now widely performed using morphological analyzers like MeCab and Chasen and their performances are usually very high level. However, they are unable to segment broken expressions or the endings of utterances that are not found in the dictionary, which often appears in lines of anime or game characters (refer to Section 2). To hinder this problem, we propose using subword units to segment lines of Japanese anime or game characters and extracting strings that occur frequently (refer to Section 3). The subword units are usually used with deep learning technologies and their robustness for out-of-vocabulary words is often noted. However, they are less interpretable than the original words because the segmentation are depending on the frequencies or occurrence probabilities rather than the meanings. In the current study, however, we show that the expressions extracted using subword units are more interpretable than those using the original words for the extractions of linguistic speech patterns of fictional characters, which is the case where many words are not listed in the dictionary using data collected from publications on the internet (refer to Section 4). We also show that the subword units are effective even though no deep learning technology is used with them. In the experiment, we weighted the subword units by TF/IDF according to gender, age, and each anime character (refer to Sections 5) and show that they are linguistic speech patterns that are specific for each feature (refer to Sections 7 and 8). Additionally, we performed a classification experiment using a support vector machine (SVM) based on linguistic speech patterns we extracted to classify the characters into a character group (refer to Sections 6) and showed that a subword unit model outperformed a conventional morphological analyzer (refer to Sections 7 and 8). Finally, we conclude our work in Section 9. This paper is an extended version of "Extraction of Linguistic Speech Patterns of Japanese Fictional Characters Using Subword Units", published in the proceedings of 10th International Conference on Natural Language Processing (NLP 2021).

## 2. RELATED WORK

Japanese does not have word delimiters between words and word boundaries in Japanese are unspecific. Therefore, there has been much research on Japanese word segmentation or morphological analysis and there are many morphological analyzers for Japanese texts like MeCab [2], Chasen, Juman++ [3], and KyTea [4], These morphological analyzers segment words with high performances but sometimes the performances decrease for the noisy texts. For Japanese word segmentation of noisy texts, Sasano et al. [5] proposed a simple approach to unknown word processing, including unknown onomatopoeia in Japanese morphological analysis. Saito et al. [6] also recommend using character-level and word-level normalization to address the morphological analysis of noisy Japanese texts. Recently, algorithms for subword unis such as Byte Pair Encoding (BPE) [7] and unigram language model [8] are proposed. They are mainly proposed for neural machine translation and usually used with deep learning technologies. We used the unigram language model for word segmentation of Japanese lines of fictional characters. There are some studies on interpretability and usability of words depending on the word segmentation for information retrieval (IR). Kwok [9] investigated and compared 1-gram, bigram, and short-word indexing for IR. Nie et al. [10] proposed the longest-matching algorithm with single characters for Chinese word segmentation for IR. In addition, there has been much research on characterization. PERSONAGE (personality generator) developed by Mairesse and Walker [11] as, the first highly parametrizable conversational language generator. They produced recognizable linguistic variation and personality, and our work also focused on each character's personality. Walker et al. [12] reported a corpus of film dialog collected and annotated for linguistic structures and character archetypes. Additionally, they conducted experiments on their character models to classify linguistic styles depending on groups such as genre, gender, directors, and film period. Miyazaki et al. [13] conducted a fundamental analysis of Japanese linguistic expressions that characterize speeches for developing a technology to characterize conversations by partially paraphrasing them. In their subsequent research, Miyazaki et al. [14] reported categories of linguistic peculiarities of Japanese fictional characters. Miyazaki et al. [15] conducted an experiment to see whether the reader can understand the characterization of a dialog agent by paraphrasing the functional part of each sentence with a probability suitable for the target character, as a way to characterize the speech and to enrich the variation of the speeches. Another study focused on Japanese sound change expressions to characterize speeches of Japanese fictional characters; they collected these expressions and classified them [16]. Additionally, Okui and Nakatsuji [17] used a pointer generating mechanism to generate various responses for a Japanese dialog system, referring to several different character responses. They learned the characterization of the responses with a small amount of data.

## 3. EXTRACTION OF LINGUISTIC SPEECH PATTERNS USING SUBWORD UNITS

Many terms not included in the dictionary such as expressions with characterization at the endings of utterances and broken expressions appear in fictional character dialogs. As a result, using existing morphological analyzers with dictionaries to segment the lines of fictional characters are challenging. Therefore, we propose using subword units for the segmentation of lines of fictional characters. The concept behind subword units is that the frequency of occurrence of a word is studied in advance, and low-frequency words are broken down into letters and smaller words. In other words, using subword units, we can treat a string with a high frequency of occurrence as a single unit, not a word in a dictionary. We used software referred to SentencePiece [18] for word segmentation of Japanese lines of fictional characters. SentencePiece learns the segment method directly from the text and segments the text into subword units. It supports BPE and unigram language model, but we employed unigram language model because it slightly outperformed BPE when they were used for machine translation.

### 3.1. Unigram Language Model

We explain the algorithm of unigram language model quoting from [8]. The unigram language model makes an assumption that each subword occurs independently, and consequently, the probability of a subword sequence $X = (x_m, \ldots, x_m)$ is formulated as the product of the subword occurrence probabilities $p(x_i)$. The most probable segmentation $X^*$ for the input sentence $X$ is obtained with the Viterbi algorithm. Because the vocabulary set $V$ is unknown, they seek to find them with the following iterative algorithm.

1. Heuristically make a reasonably big seed vocabulary $V$ from the training corpus.

2. Repeat the following steps until $|V|$ reaches a desired vocabulary size.

    (a) Fixing the set of vocabulary, optimize $p(x)$ with the EM algorithm.

    (b) Compute the $loss_i$ for each subword $x_i$, where $loss_i$ represents how likely the likelihood is reduced when the subword $x_i$ is removed from the current vocabulary.

    (c) Sort the symbols by $loss_i$ and keep top $\eta$% of subwords.

Unigram language model is a method whose objective function is maximization of log likelihood of "X" .

### 3.2. Procedures

We extracted linguistic speech patterns that characterize the lines as follows:

1. Collect lines of fictional characters,

2. Segment the lines into subword units using SentencePiece, and

3. Weighted the subword units using TF/IDF values and obtain the top ten subword units.

In addition to the extraction experiments, we conducted classification experiments of characters. Finally, we compared the results of the method using SentencePiece with that of one of the de facto standard morphological analyzers for Japanese, MeCab. We used ipadic for Japanese dictionary of MeCab.

## 4. DATA

We collected dialogs of 103 characters from 20 publications on the internet. They are, Anohana: The Flower We Saw That Day, Den-noh Coil, Dragon Quest IV-VIII, Neon Genesis Evangelion, Mobile Suit Gundam, Howl's Moving Castle, Hyouka, Kaguya-sama: Love Is War, Kemono Friends, Harem Days, Whisper of the Heart, Laputa: Castle in the Sky, Spirited Away, Symphogear, My Neighbour Totoro, and The Promised Neverland. This corpus of dialogs is referred as to the "Character Corpus." The following three methods were used for the collection.

1. They were collected from a compilation site of anime and game dialog on the internet.

2. They were collected from anime video sites.

3. It was converted from manga e-books using a text detection application.

Priority was given to characters with many lines while choosing character in the work. Furthermore, since it was assumed that the majority of the main characters would be mostly classified as boys, girls or younger men or women, we aggressively collected child and older characters with a significant number of dialogs during the selection process. Because we have classification experiments according to age, characters whose ages change drastically during the story have been removed. An example of this is Sophie from Howl's Moving Castle. She changed from 18 to 90 years old in the movie. We also eliminated characters with extremely low amounts of dialog. The minimum, maximum and average numbers of lines of a character are respectively 92, 6,797, and 1,187.17.

## 5. EXPERIMENTS OF LINGUISTIC SPEECH PATTERN EXTRACTION

The procedure of linguistic speech pattern extraction by SentencePiece is as follows. First, we develop a segmentation model by applying SentencePiece to each character's dialog. Notably, we apply SentencePiece to sub-corpus of each character rather than the entire corpus. This is because the way of talking varies according to each character. The following formula calculates the maximum number of subword units:

$$\text{Vocabulary\_Size} = \text{Basic\_VS} * \left(\frac{L}{l}\right)^{\frac{1}{5}} \quad (1)$$

where, $l$ denotes the number of letters of each character's lines, $L$ denotes the total number of letters of lines of all characters, and $\text{Basic\_VS}$ denotes basic vocabulary size. We set $\text{Basic\_VS}$ to 3,000. Simultaneously with the creation of the model, a word list from the vocab file was also constructed. We delete from the word list subword units that consist of a single Chinese character except for the first-person singular(僕, 私, 俺) because we believed that they would not express a characterization. We also deleted 1/5 of the subword units with less emission logarithmic establishment, which is a measure of a subword unit's occurrence probability. As a result, the number of words was 9114. For the next step, we segment the character corpus using the segmentation model we created. The word lists and segmented character corpus were used to obtain the TF/IDF value, which was calculated using the following formula

$$tf(t, d) = \frac{n_{(t, d)}}{\sum_{s \in d} n_{(s, d)}} \tag{2}$$

where, $tf(t, d)$ denotes the term frequency of a subword unit $t$ in document $d$, $n_{(t, d)}$ denotes the number of occurrences of a subword unit $t$ in document $d$, $\sum_{s \in d} n_{(s, d)}$ denotes the sum of the number of occurrences of all subword units in the document $d$.

$$idf(t) = \log \frac{N}{df(t)} \tag{3}$$

where, $idf(t)$ denotes the inverse document frequency of a subword unit $t$, $N$ denotes total number of documents, $df(t)$ denotes number of documents in which a subword unit $t$ occurred.

$$TF/IDF = tf(t, d) * idf(t) \tag{4}$$

We extracted linguistic speech patterns that characterize lines of gender, ages, and characters using TF/IDF value. We considered the lines of all characters of one gender as one document, and the lines of all character of the opposite gender as another document when calculating the TF/IDF value for a gender. A summary of the experimental procedure is shown in Figure 1.

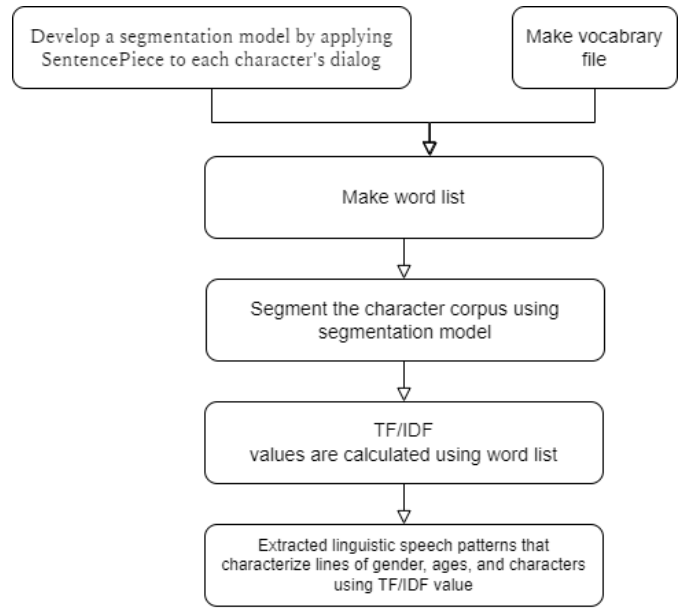

Figure1.  Flowchart of the extraction experiment procedure

## 6. CLASSIFICATION EXPERIMENT

We performed a classification experiment to evaluate the extracted linguistic speech patterns using a SVM. The obtained TF/IDF values were used as inputs to the SVM to classify the characters into groups categorized by gender and age. The characters were first divided into three categories: children, adults, and seniors. Children and adults were further divided into two categories: male and female whereas seniors have only one group because we had few characters of the ages. As a result, we used five groups: boys, girls, men, women, and seniors. The numbers of the character according to the group are shown in Table1. The group classification was performed based on the character's characterization and not on their actual age or gender because the profiles of fictional characters are sometimes extraordinary. The bias in the amount of data for each category is affected by the bias in characters; Japanese anime and games we collected have a few children and senior characters. The experiment was conducted using five-fold cross-validation. Sklearn was used as a library in this experiment. The computational complexity of SVM in sklearn varies between O (number of dimensionalities * number of data ^2) and O (number of dimensionality * number of samples ^3), depending on how efficiently the cache is used.

Table 1. Amount of data for Classification Experiment.

| Boys | Girls | Men | Women | Senior |
|---|---|---|---|---|
| 6 | 8 | 40 | 41 | 7 |

## 7. RESULTS

The linguistic speech patterns with the top 10 TF/IDF values are shown in Tables 2-7. In the tables, E represents ending, and F means first-person singular. Tables 2 and 3 list the linguistic speech patterns with gender characterization, and Tables 4 and 5 show those with age characterization. In these tables, Italic means that the pattern is specific for each characterization of fictional characters. Some of the character-specific linguistic speech patterns are also shown as example results in Tables 6 and 7. The example characters are Emma from the anime "The Promised Neverland," Shinji from the anime "Neon Genesis Evangelion," and Yangus from the game "Dragon Quest VIII." For the experiment of characters, we had a questionnaire to evaluate the linguistic speech patterns. Eight native Japanese speakers were asked if each linguistic speech pattern seems specific for the character. Five people are men, and three were women, and seven people are in their 20's, and one person is in her 30's. They were also asked if they knew each anime or game that the character appears. Tables 8 and 9 summarize the results of the questionnaire. Finally, the results of the classification experiment are shown in Table 10.

Table 2. Linguistic Speech Patterns with Gender Characterization Retrieved by SentencePiece. E represents ending and F denotes the first-person singular. Italic means that the pattern is specific for the people of specific genders as lines of fictional characters.

| Male | | | Female | | |
| --- | --- | --- | --- | --- | --- |
| **Patterns** | **Sounds** | **Notes** | **Patterns** | **Sounds** | **Notes** |
| ですね | desune | Polite E | わね | wane | *Feminine E* |
| でござる | degozaru | *Samurai E* | かしら | kashira | *Feminine E* |
| だぜ | daze | *Masculine E* | のかしら | nokashira | *Feminine E* |
| でござるな | degozaruna | *Samurai E* | だわ | dawa | *Feminine E* |
| アルス | Arusu | Name | よね | yone | *E* |
| だな | dana | *Masculine E* | のね | none | *Feminine E* |
| なあ | naa | *Old buddy* | ないわ | naiwa | *Feminine E* |
| ますね | masune | Polite E | わよ | wayo | *Feminine E* |
| でがすよ | degasuyo | *Dialect E* | ないわね | naiwane | *Feminine E* |
| でござるよ | degozaruyo | *Samurai E* | アルス | Arusu | Name |

Table 3. Linguistic Speech Patterns with Gender Characterization Retrieved by MeCab. E represents ending and F denotes the first-person singular. Italic means that the pattern is specific for the people of specific genders as lines of fictional characters.

| Male | | | Female | | |
| --- | --- | --- | --- | --- | --- |
| **Patterns** | **Sounds** | **Notes** | **Patterns** | **Sounds** | **Notes** |
| ござる | gozaru | *Samurai E* | あたし | atashi | *F for girls* |
| ざる | zaru | Error | かしら | kashira | *Feminine E* |
| 俺 | ore | *F for male* | アルス | Arusu | Name |
| アルス | Arusu | Name | 私 | watashi | *F* |
| オイラ | oira | *F for boys* | . | . | Mark |
| げす | gesu | *Dialect E* | しら | shira | Error |
| . | . | Mark | リュカ | Ryuka | Name |
| 僕 | boku | *F for boys* | たし | tashi | Error |
| ウィル | Will | Name | ましょ | masyo | *Femminine E* |
| 俺 | ore | *F for male* | ウィル | Will | Name |

Table 4. Linguistic Speech Patterns with Age Characterization Retrieved by SentencePiece. E represents ending and F denotes the first-person singular. Italic means that the pattern is specific for the people of specific ages as lines of fictional characters.

| Children | | | Adults | | | Seniors | | |
| --- | --- | --- | --- | --- | --- | --- | --- | --- |
| **Patterns** | **Sounds** | **Notes** | **Patterns** | **Sounds** | **Notes** | **Patterns** | **Sounds** | **Notes** |
| なあ | naa | Old buddy | ですね | deshune | Polite E | でござる | degozaru | *Samurai E* |
| アルス | Arusu | Name | わね | wane | *Feminine E* | でござるな | degozaruna | *Samurai E* |
| お父さん | otosan | *Dad* | これ | kore | This | でござるよ | degozaruyo | *Samurai E* |
| オイラ | oira | *F for boys* | です | deshu | Polite E | でござるか | degozaruka | *Samurai E* |
| だよ | dayo | E | だな | dana | Masculine E | アルス殿 | Arushudono | Sir Arusu |
| いっぱい | ippai | *Many* | かしら | kashira | *Feminine E* | 殿 | dono | Sir |
| だぞ | dazo | *Boyish E* | なんて | nante | Exclamatory how | るでござるよ | rudegozaruyo | *Samurai E* |
| てる | teru | E | アルス | Arusu | Name | とは | towa | C.f. with |
| るの | runo | *Feminine E* | どこ | doko | Where | るでござる | rudegozaru | *Samurai E* |
| だね | dane | E | さん | san | *title* | わし | washi | *F for old men* |

Table 5. Linguistic Speech Patterns with Age Characterization Retrieved by MeCab. E represents ending and F denotes the first-person singular. Italic means that the pattern is specific for the people of specific ages as lines of fictional characters.

| Children | | | Adults | | | Seniors | | |
|---|---|---|---|---|---|---|---|---|
| **Patterns** | **Sounds** | **Notes** | **Patterns** | **Sounds** | **Notes** | **Patterns** | **Sounds** | **Notes** |
| オイラ | oira | *F for boys* | 俺 | ore | *F for male* | ござろ | gozaro | *Samurai E* |
| 僕 | boku | *F for boys* | ウィル | Will | Name | ござっ | goza | *Samurai suffix* |
| おっちゃん | ochan | *Pops* | リュカ | Ryuka | Name | など | nado | Such as |
| ゃっ | ya | Error | げす | gesu | Dialect | フム | humu | *Hm-hum* |
| ちゃっ | cha | Error | アムロ | Amuro | Name | うむ | umu | *Hmmm* |
| ちゃう | chau | *End up -ing* | ひすい | hisui | Name | やはり | yahari | As expected |
| オラ | ora | *F for boys* | ゃっ | ya | Error | サントハイム | santohaimu | Name |
| うわ | uwa | Wow | ちゃっ | cha | Error | いかん | ikan | *No for old men* |
| じんた | jinta | Error | アニキ | aniki | Bro | むう | muu | *Hmmm* |
| オッチャン | ochan | *Pops* | ドルマゲス | Dhoulmagus | name | ふむ | humu | *Hm-hum* |

Table 6. Character-specific Linguistic Speech Patterns Retrieved by SentencePiece. E represents ending and F denotes the first-person singular.

| Emma | | | Shinji | | | Yangus | | |
|---|---|---|---|---|---|---|---|---|
| **Patterns** | **Sounds** | **Notes** | **Patterns** | **Sounds** | **Notes** | **Patterns** | **Sounds** | **Notes** |
| てる | teru | E | ですか | desuka | Polite E | でがすよ | degasuyo | Dialect |
| にも | nimo | And | ミサトさん | Misatosan | Name with title | でがす | degasu | Dialect |
| ってこと | ttekoto | That means | 僕は | bokuha | I am (F for men) | でげす | degesu | Dialect |
| ちょ | cho | Wait | ないよ | naiyo | there isn't | でげすよ | degesuyo | Dialect |
| いいよ | iiyo | OK | 父さん | tosan | Dad | でがすね | degasune | Dialect |
| 嫌だ | iyada | No | るんだ | runda | E | おっさん | ossan | Pops |
| の手 | note | Hand of | 僕 | boku | F for men | かい | kai | E |
| 私たちの | watashitachino | Our | だよ | dayo | E | んでがす | ndegasu | Dialect |
| 信じ | shinji | Believe | 綾波 | Ayanami | Name | アッシは | asshiha | I am for men |
| もし | moshi | If | んですか | ndesuka | E for question | アッシら | asshira | We for men |

Table 7. Character-specific Linguistic Speech Patterns Retrieved by MeCab. E represents ending and F denotes the first-person singular.

| Emma | | | Shinji | | | Yangus | | |
|---|---|---|---|---|---|---|---|---|
| **Patterns** | **Sounds** | **Notes** | **Patterns** | **Sounds** | **Notes** | **Patterns** | **Sounds** | **Notes** |
| 私 | watashi | F | 僕 | boku | F for men | げす | gesu | Dialect |
| レイ | Rei | Name | ミ | mi | Error | がす | gasu | Dialect |
| ノーマン | Noman | Name | サト | sato | Error | アッ | a | Error |
| マン | man | Error | 父さん | tosan | Dad | アッシ | asshi | F for men |
| 思う | omou | Think | さん | san | Title | すね | sune | E |
| うん | un | Yes | うわ | uwa | Wow | すか | suka | E for question |
| 近寄っ | chikayo | Draw near | スカ | suka | Error | やしょ | yasho | Dialect |
| 折れ | ore | Be folded | アスカ | Asuka | Name | おっさん | ossan | Pops |
| 寄っ | yo | Draw near | トウジ | Touji | Name | 姉ちゃん | nechan | Sis |
| そっ | so | Error | 僕ら | bokura | We for men | ダンナ | danna | Master |

Table 8. Number of People Who Think the Linguistic Speech Pattern Extracted by SentencePiece is specific for the Character and Its Percentages. W/ represents with knowledge of the anime or game and w/o indicates without knowledge. People represents number of people with and without knowledge of the anime or game.

| Emma | | | Shinji | | | Yangus | | |
|---|---|---|---|---|---|---|---|---|
| **Patterns** | **w/** | **w/o** | **Patterns** | **w/** | **w/o** | **Patterns** | **w/** | **w/o** |
| **People** | **4** | **4** | **People** | **7** | **1** | **People** | **3** | **5** |
| てる | 0 | 0 | ですか | 0 | 0 | でがすよ | 3 | 4 |
| にも | 0 | 0 | ミサトさん | 1 | 0 | でがす | 3 | 4 |
| ってこと | 0 | 0 | 僕は | 3 | 1 | でげす | 3 | 4 |
| ちょ | 0 | 0 | ないよ | 0 | 0 | でげすよ | 3 | 4 |
| いいよ | 1 | 0 | 父さん | 3 | 1 | でがすね | 3 | 4 |
| 嫌だ | 3 | 0 | るんだ | 1 | 1 | おっさん | 1 | 2 |
| の手 | 0 | 0 | 僕 | 2 | 1 | かい | 1 | 2 |
| 私たちの | 4 | 1 | だよ | 0 | 0 | んでがす | 3 | 4 |
| 信じ | 3 | 0 | 綾波 | 4 | 0 | アッシは | 3 | 3 |
| もし | 0 | 0 | んですか | 0 | 0 | アッシら | 3 | 3 |
| Total | 11 | 1 | Total | 14 | 4 | Total | 26 | 34 |
| Percent | 27.50% | 2.50% | Percent | 20.00% | 40.00% | Percent | 86.67% | 68.00% |
| Avarage | 15.00% | | Avarage | 22.50% | | Avarage | 75.00% | |

Table 9. Number of People Who Think the Linguistic Speech Pattern Extracted by MeCab is specific for the Character and Its Percentages. W/ represents with knowledge of the anime or game and w/o indicates without knowledge. People represents number of people with and without knowledge of the anime or game.

| Emma | | | Shinji | | | Yangus | | |
|---|---|---|---|---|---|---|---|---|
| Patterns | w/ | w/o | Patterns | w/ | w/o | Patterns | w/ | w/o |
| People | 4 | 4 | People | 7 | 1 | People | 3 | 5 |
| 私 | 0 | 0 | 僕 | 2 | 1 | げす | 3 | 4 |
| レイ | 1 | 0 | ミ | 0 | 0 | がす | 3 | 4 |
| ノーマン | 1 | 0 | サト | 0 | 0 | アッ | 0 | 0 |
| マン | 0 | 0 | 父さん | 3 | 1 | アッシ | 3 | 3 |
| 思う | 0 | 0 | さん | 0 | 0 | すね | 2 | 1 |
| うん | 0 | 0 | うわ | 0 | 0 | すか | 2 | 1 |
| 近寄っ | 0 | 0 | スカ | 0 | 0 | やしょ | 2 | 1 |
| 折れ | 0 | 0 | アスカ | 4 | 0 | おっさん | 1 | 2 |
| 寄っ | 0 | 0 | トウジ | 0 | 0 | 姉ちゃん | 0 | 3 |
| そっ | 0 | 0 | 僕ら | 1 | 1 | ダンナ | 1 | 3 |
| Total | 2 | 0 | Total | 10 | 3 | Total | 17 | 22 |
| Percent | 5.00% | 0.00% | Percent | 14.29% | 30.00% | Percent | 56.67% | 44.00% |
| Avarage | 2.50% | | Avarage | 16.25% | | Avarage | 48.75% | |

Table 10. Results of Classification Experiment

| SentencePiece | MeCab |
|---|---|
| 0.627 | 0.451 |

## 8. DISCUSSION

### 8.1. Discussion of the extraction experiment

Tables 2-7 shows that regardless of whether the SentencePiece or MeCab model is used, many endings of utterances and first-person singulars are extracted as specific linguistic speech patterns. We believe that they substantially characterize Japanese dialog. Many personal names are also extracted, although they are not linguistic speech patterns, because they often appeared in the lines of characters. MeCab found 13 error expressions whereas the SentencePiece model found none. Here, an error means the expression has no meaning due to a segmentation error. This result indicates that a conventional morphological analyzer sometimes fails to segment unusual sentences such as lines of fictional characters. Furthermore, we can observe from the tables that the SentencePiece model can obtain linguistic speech patterns that consist of many words. For example, "desune" consists of "desu" and "ne" and "wane" consists of "wa" and "ne." The SentencePiece model could retrieve these linguistic speech patterns because it used subword units. In Mecab, on the other hand, the word "アッシ, asshi" was split into "アッ, a" and "シ, shi". "アッシ, Asshi" is the first person singular of Yangus, and it is one word,

splitting it would cause an error. As you can see, many errors occurred in the Mecab extraction process when splitting words that were not in the dictionary.

Furthermore, the SentencePiece model retrieved seven masculine and nine feminine linguistic speech patterns for the gender experiment, whereas MeCab retrieved six masculine and four feminine linguistic speech patterns. SentencePiece extracted a large number of endings that represent gender characteristics. For example, "だぜ, daze" and "だな, dana" are often used by men as endings of utterances. And "わね, wane" and "かしら, kashira," are thought to be found in the endings of female speeches. Although not listed in table2, the first person singular was also taken to indicate gender characteristics. For example "私, watashi" is a first person singular mostly for women and "俺, ore" is that for men in fictions. In the extraction using SentencePiece, many feature words were obtained in addition to the top 10 words. Also, MeCab extracted symbols and words that did not express the meaning and did not capture the features as well as SentencePiece.

For age experiment, the SentencePiece model obtained six, two, and seven linguistic speech patterns that are children, adults, and seniors, respectively, whereas MeCab retrieved six, one, seven linguistic speech patterns. For example, "お父さん, Otosan" means dad is a commonly used word, but we thought this was indicative of age characteristics, because we could imagine it being used in situations where children call their parent. Although "いっぱい, ippai" means many, is also a commonly used word, there are other paraphrases such as "たくさん, takusan" and "多く, ooku", but this is a particularly childish expression, so we considered it as a characteristic word. The honorific "さん, san" is not often used by children in anime and games, and it is not used so often by older people because their position and age are often higher than those of other characters. Also, if anything, children in anime and games tend to use "ちゃん, chan" instead of "さん, san". The word in table 5 "おっちゃん, ochan," which is a characteristic word for children, is not often used by adults because it can be seen as a rude expression. The first person singular for children, "僕, boku" is used by men in real life, but in anime and games, it is often used by child characters. "オイラ, Oira" and "オラ, ora" is also used by child characters in anime and games, but not in real life. The characteristic words of the senior generation, such as "ふむ, hmm" and "うむ, um", are often used as a gesture of thinking by the senior generation. For the experiment of ages, the difference between the two models was smaller than that of gender. We believe that the systems could not extract linguistic speech patterns specific for adults because their talking way is considered normal. In the table2-5 that shows the results of the experiment of extracting linguistic speech patterns for age and gender, SentencePiece extracted mostly ending of utterances, while MeCab extracted not only ending of utterances but also first person singulars and broken expressions such as "おっちゃん, ochan".

In the tables of the experiments, it may seem that Mecab extracted more kinds of feature words than SentencePiece. However, most of the feature words extracted by Mecab could also be extracted by SentencePiece. The examples of the linguistic speech patterns extracted by SentencePiece that could not be showed are shown in Table 11-12.

Next, let us discuss the experiment of each character. This is more difficult than the discussion of gender or age because the knowledge of the character can affect the results. Therefore, we had a questionnaire for eight people. Tables 8 and 9 show that the SentencePiece model always obtains more character-specific linguistic speech patterns than MeCab for every character. The knowledge of the characters did not affect this result. However, the people with knowledge considerably feel that the linguistic speech patterns are specific for Emma, but the people without knowledge feel they are not so much. According to English Wikipedia, "The bright and cheerful Emma is an 11-year-old orphan living in Grace Field House, a self-contained orphanage housing her and 37 other orphans." We believe that people without knowledge tend to think she is an adult woman because Emma is a female name. People without knowledge could think that the extracted patterns are not character-specific because they include no feminine patterns. Additionally, according to the Dragon Quest Wiki, "Yangus is a character in Dragon Quest VIII who accompanies the Hero on his missions." and "He serves as a powerful tank character over the course of the game." As for Yangus, the people with knowledge feel that the linguistic speech patterns are more character-specific again. Moreover, the percentage where people think they are character-specific is the highest among the three characters. We believe that this is because Yangus speak a dialect-originated and specific language. According to English Wikipedia, "Shinji is a dependent and introverted boy with few friends, reluctant unable to communicate with other people and frightened by contact with strangers." Although there could be a bias because only one person did not know him, the person who did not know felt that the extracted patterns were more character-specific. These results indicate that the extracted expressions using subword units are more interpretable linguistic speech patterns than those using words. Also, because of Emma and Shinji spoke in a way that real people also use, there were few words that could be extracted as feature words. For characters like Yangus, who spoke in a way unique to anime and games, more feature words could be extracted.

The system was able to successfully extract linguistic speech patterns, but it was not perfect. In addition to linguistic speech patterns, many proper nouns and common nouns that appear frequently in the work were extracted because this system extracts strings that appear frequently. In particular, it is difficult to remove common nouns automatically. This is because some of the common nouns extracted can be considered as linguistic speech patterns, such as "お父さん, otosan", which means father, and some cannot be considered as them, such as "お城, Oshiro", which means castle. As a solution to these problems, the following methods can be considered. We think that proper nouns can be removed by referring to word lists and dictionaries. Words that are not in the dictionary or that are written as proper nouns in the dictionary are proper nouns and can be removed. We considered that there are two types of common nouns: nouns that describe the characteristics of characters and nouns that often appear in the story. Of these, we want to remove the nouns that appear frequently in the story, since they do not represent the characteristics of the characters. It may be possible to remove them by extracting words that appear frequently in each work using the same method as in the extraction experiment. The other problem is that there are few characters from the seniors and the children. The solution to this problem is very difficult.

### 8.2. Discussion of the classification experiment

The classification results also showed that the SentencePiece model outperformed MeCab for the classification of character groups. Additionally, it indicates that the patterns are more specific for each character group feature. Notably, the subword units are proposed for deep learning technologies but our classification did not use any of them. The experiments showed that the subword units are effective when no deep learning technologies are used.

Table 11. Examples of linguistic speech patterns with gender characterization extracted by SentencePiece from the top 11. E represents ending and F denotes the first-person singular.

| Male | | | Female | | |
|---|---|---|---|---|---|
| Patterns | Sounds | Notes | Patterns | Sounds | Notes |
| んだな | ndana | Masculine E | あたし | atashi | F for Female |
| ですな | desuna | Masculine E | ですわ | desuwa | Lady E |
| オレ | ore | F for male | そうね | sone | Feminine E |
| オイラ | oira | F for boys | 私たち | watashi-tachi | We |
| 行こうぜ | ikoze | Let's go | なさいよ | nasaiyo | Do it |
| ちまった | chimatta | Did it | あたしたち | atashi-tachi | We |
| ないぜ | naize | No | あんた | anta | You |
| 兄貴 | aniki | Bro | ないわよ | naiwayo | No |
| 僕 | boku | F for male | あなた | anata | You |
| お前 | omae | You | お父さま | otosama | Father |

Table 12. Examples of linguistic speech patterns with age characterization extracted by SentencePiece from the top 11. E represents ending and F denotes the first-person singular.

| Children | | | Adults | | | Seniors | | |
|---|---|---|---|---|---|---|---|---|
| Patterns | Sounds | Notes | Patterns | Sounds | Notes | Patterns | Sounds | Notes |
| ボク | boku | F for boys | のかしら | nokashira | E for female | ないでござる | naide gozaru | No |
| お母さん | okasan | Mam | わよ | wayo | E for female | わい | wai | F for seniors |
| ねー | ne | Hey | ません | masen | polite E | わしら | washira | We |
| もん | mon | E for children | なあ | naa | Hey | まい | mai | Not |

## 9. CONCLUSIONS

In this study, we proposed using subword units to segment dialogs of fictional characters. The experiments revealed that subword units weighted with TF/IDF values are character-specific linguistic speech patterns, that cannot be obtained from existing morphological analyzers using dictionaries. They also showed that the linguistic speech patterns retrieved using SentencePiece are more specific for gender, age, and each character. It indicates that the extracted expressions using subword units are more interpretable than those using words. We discussed the differences between the extracted linguistic speech patterns retrieved using SentencePiece and famous morphological analyzer, MeCab. We also conducted an experiment that classifies the characters into character groups using the extracted linguistic speech patterns as features, and the classification SentencePiece model's accuracy was compared to the case where MeCab was used to segment the dialogs. We showed that subword units are effective even though no deep learning technologies are used with them. In the future, we would like to consider parts of speech when segment terms. Also, we are interested in research to generate sentences with characterization using the linguistic speech patterns extracted in this study.

# REFERENCES

[1] Satoshi Kinsui, (2017) *Virtual Japanese : Enigmas of Role Language*, Osaka University Press.

[2] Taku Kudo and Kaoru Yamamoto and Yuji Matsumoto, (2004) *Applying Conditional Random Fields to Japanese Morphological Analysis*, the Proceedings of EMNLP 2004, pp230-237.

[3] Hajime Morita and Daisuke Kawahara and Sadao Kurohashi, (2015) *Morphological Analysis for Unsegmented Languages using Recurrent Neural Network Language Model*, the Proceedings of EMNLP 2015, pp 2292-2297.

[4] Graham Neubig and Yosuke Nakata and Shinsuke Mori, (2011) *Pointwise Prediction for Robust, Adaptable Japanese Morphological Analysis*, the Proceedings of ACL-HLT 2011, pp 529-533.

[5] Ryohei Sasano and Sadao Kurohashi and Manabu Okumura, (2013) *A simple approach to unknown word processing in japanese morphological analysis*, the Proceedings of the Sixth International Joint Conference on Natural Language Processing, pp 162-170.

[6] Itsumi Saito and Kugatsu Sadamitsu and Hisako Asano and Yoshihiro Matsuo, (2014) *Morphological Analysis for Japanese Noisy Text Based on Character-level and Word-level Normalization*, the Proceedings of COLING 2014, pp 1773-1782.

[7] Rico Sennrich and Barry Haddow and Alexandra Birch, (2016) *Neural Machine Translation of Rare Words with Subword Units*, the Proceedings of the 54th ACL, pp1715-1725.

[8] Taku Kudo, (2018) *Subword Regularization: Improving Neural Network Translation Models with Multiple Subword Candidates*, the Proceedings of ACL 2018, pp 66-75.

[9] K.L. Kwok, (1997) *Comparing Representations in Chinese Information Retrieval*, the Proceedings of the 20th annual international ACM SIGIR conference on Research and development in information retrieval, pp34-41.

[10] Jian-Yun Nie and Jiangfeng Gao and Jian Zhang and Ming Zhou, (2000) *On the Use of Words and N-grams for Chinese Information Retrieval*, the Proceedings of the Fifth International Workshop on Information Retrieval with Asian Languages, pp141-148.

[11] François Mairesse and Marilyn Walker, (2007) *PERSONAGE: Personality Generation for Dialogue*, the Proceedings of ACL 2007, pp496-503.

[12] Marilyn A. Walker and Grace I. Lin and Jennifer E. Sawyer, (2012) *An Annotated Corpus of Film Dialogue for Learning and Characterizing Character Style*, the Proceedings of LREC 2012, pp1373–1378.

[13] Chiaki Miyazaki and Toru Hirano and Ryuichiro Higashinaka and Toshiro Makino and Yoshihiro Matsuo and Satoshi Sato, (2014) *Basic Analysis of Linguistic Peculiarities that Contribute Characterization of Dialogue Agent*, the Proceedings of NLP2014, pp232-235(In Japanese).

[14] Chiaki Miyazaki and Toru Hirano and Ryuichiro Higashinaka and Yoshihiro Matsuo, (2016) *Towards an Entertaining Natural Language Generation System: Linguistic Peculiarities of Japanese Fictional Characters*, the Proceedings of SIGDIAL 2016, pp319–328.

[15] Chiaki Miyazaki and Toru Hiranoand Ryuichiro Higashinaka and Toshiro Makino and Yoshihiro Matsuo, (2015) *Automatic conversion of sentence-end expressions for utterance characterization of dialogue systems*, the Proceedings of PACLIC 2015, pp307–314.

[16] Chiaki Miyazaki and Satoshi Sato, (2019) *Classification of Phonological Changes Reflected in Text: Toward a Characterization of Written Utterances*, Journal of Natural Language Processing, Vol. 26, No.2, pp407-440(In Japanese).

[17] Sohei Okui and Makoto Nakatsuji, (2020) *Evaluating response generation for character by pointer-generator-mechanism*, Proceedings of the 34th Annual Conference of the Japanese Society for Artificial Intelligence, pp1I4-GS-2-01(In Japanese).

[18] Taku Kudo and John Richardson, (2018) *SentencePiece: A simple and language independent subword tokenizer and detokenizer for Neural Text Processing*, the Proceedings of EMNLP 2018, pp66-71.

**Authors**

Short Biography

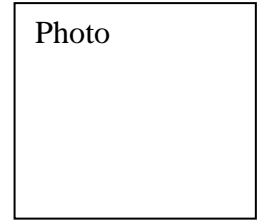